\documentclass[runningheads]{llncs}
\usepackage[T1]{fontenc}
\usepackage{graphicx}
\usepackage{multirow}
\usepackage{xcolor}
\usepackage{colortbl}
\usepackage{booktabs}
\usepackage{ulem}
\usepackage{pifont}
\usepackage{amsfonts,amssymb,amsmath}
\usepackage{dsfont}
\usepackage{url}
\usepackage{marvosym}

\begin{document}
\title{Generalizing to Unseen Domains in Diabetic Retinopathy with Disentangled Representations}
\titlerunning{Generalizing to Unseen Domains in Diabetic Retinopathy with Disentangled Representations}
%
\author{Peng Xia\inst{1,2}\thanks{Equal Contribution} \and
Ming Hu\inst{1}$^{\star}$ \and
Feilong Tang\inst{1} \and
Wenxue Li\inst{1} \and
Wenhao Zheng\inst{2} \and \\
Lie Ju\inst{1} \and 
Peibo Duan\inst{1} \and
Huaxiu Yao\inst{2}\Letter \and
Zongyuan Ge\inst{1}\Letter
}
%
\authorrunning{P. Xia et al.}
\institute{Monash University, Melbourne, Victoria, Australia \and UNC-Chapel Hill, Chapel Hill, NC, USA\\
\email{richard.peng.xia@gmail.com,huaxiu@cs.unc.edu,zongyuan.ge@monash.edu}
}
\maketitle              
\begin{abstract}
Diabetic Retinopathy (DR), induced by diabetes, poses a significant risk of visual impairment. Accurate and effective grading of DR aids in the treatment of this condition. Yet existing models experience notable performance degradation on unseen domains due to domain shifts. Previous methods address this issue by simulating domain style through simple visual transformation and mitigating domain noise via learning robust representations. 
However, domain shifts encompass more than image styles. They overlook biases caused by implicit factors such as ethnicity, age, and diagnostic criteria. In our work, we propose a novel framework where representations of paired data from different domains are decoupled into semantic features and domain noise. The resulting augmented representation comprises original retinal semantics and domain noise from other domains, aiming to generate enhanced representations aligned with real-world clinical needs, incorporating rich information from diverse domains. Subsequently, to improve the robustness of the decoupled representations, class and domain prototypes are employed to interpolate the disentangled representations while data-aware weights are designed to focus on rare classes and domains. Finally, we devise a robust pixel-level semantic alignment loss to align retinal semantics decoupled from features, maintaining a balance between intra-class diversity and dense class features.
Experimental results on multiple benchmarks demonstrate the effectiveness of our method on unseen domains.
The code implementations are accessible on \url{https://github.com/richard-peng-xia/DECO}.

\keywords{Diabetic Retinopathy  \and Domain Generalization \and Disentangled Representations.}
\end{abstract}

\section{Introduction}
Diabetic Retinopathy (DR) is a diabetes-induced ocular disorder that affects the retina, which epitomizes one of the foremost causes of blindness~\cite{wykoff2021risk}. 
Typically, the diagnosis of DR is based on the presence of several key lesions, namely microaneurysms, hemorrhages, soft or hard exudates, hemorrhages, and cotton wool spots. Therefore, the grading of DR usually includes five categories: no DR, mild DR, moderate DR, severe DR, and proliferative DR~\cite{sebastian2023survey}.
\begin{figure}[t]
    \centering
    \includegraphics[width=0.9\linewidth]{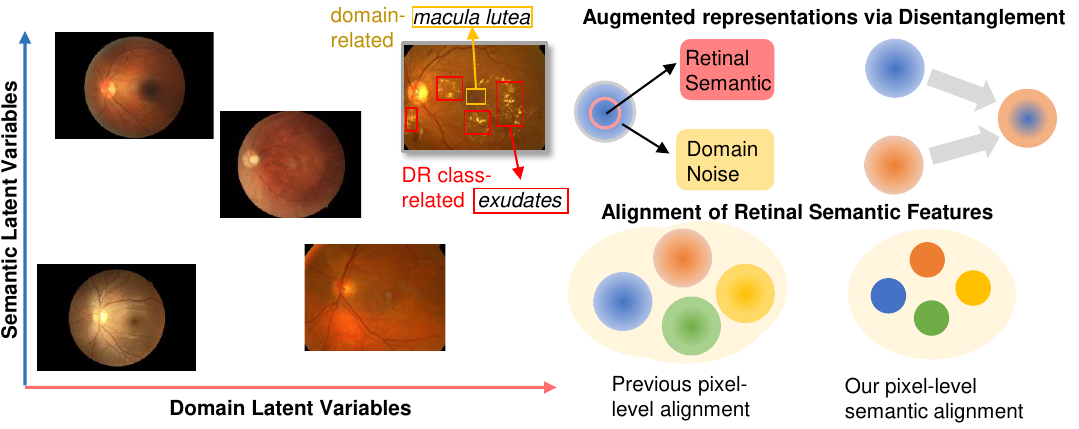}
    \caption{\textit{(Left)} An example of fundus-based domain variances. The horizontal distance represents domain differences, and the vertical distance denotes DR category differences. \textit{(Right)} The motivations of our approach. Firstly, while the augmentation methods are simple visual transformations, we consider more feature-level class-agnostic latent noise, such as macular degeneration caused by age. Additionally, existing pixel-level alignment may act on features containing domain bias, replacing original features with decoupled semantic features to alleviate domain noise.} 
    \label{fig:intro}
\end{figure}

Although deep learning methods have achieved promising results in grading DR~\cite{che2022learning,he2020cabnet,li2019diagnostic,liu2020green}, which simplifies the diagnostic process and reduces the demand for trained ophthalmologists, one major challenge they face in practical clinical applications is domain shifts~\cite{atwany2022drgen,che2023towards,che2022learning,chokuwa2023generalizing}, meaning there are some visual biases between training and testing data caused by factors such as imaging conditions or the ethnic of the population, as shown in Figure~\ref{fig:intro}. This leads to a decline in models' performance when these models are applied to new data from unseen domains, which is known as Domain Generalization (DG)~\cite{yang2022tally,yao2022improving,zhou2022domain,wu2024iterative}.

Previous efforts have sought to learn robust domain-invariant representations, such as through flatness and regularization~\cite{atwany2022drgen,yao2023improving}, and variational autoencoders~\cite{chokuwa2023generalizing}. There are works from the perspective of image augmentation~\cite{che2023towards,yao2022improving,yao2022c}, using visual transformation or image degradation to simulate domain shifts. However, generalized features are likely affected by domain biases unrelated to DR categories, such as macular degeneration in elderly retinas and ethnic variations in retinal structure. This renders current data augmentation schemes and direct feature learning ineffective when confronted with new unseen domains. Additionally, there are works proposing pixel-level supervised losses to learn diverse intra-class features~\cite{che2023towards}, but the targets themselves contain domain noise, which makes the representations less robust.
\par Considering these drawbacks, we propose to \textbf{D}ecouple the r\textbf{E}presentations of semanti\textbf{C} features from d\textbf{O}main features to reduce domain bias, which we call \textbf{DECO}. Specifically, features are decoupled into representations of semantic features (DR-related retinal semantics, \textit{e.g.}, microaneurysms, hemorrhages, hard/soft exudates) and domain features (domain noise, \textit{e.g.}, explicit noise in image style and implicit noise stemming from age, gender, and ethnicity, etc.). To mitigate the impact of domain shifts, we average the domain information over examples of the same class and construct class prototype representations. Then we linearly interpolate each semantic representation using the corresponding class prototype. Similarly, we interpolate domain representations with class-agnostic domain factors to improve training stability and remove noise. During the insertion process, we design data-aware weights to focus on rare classes and domains. By utilizing a set of data from different domains, new representations containing the semantic features of one example and domain features from another can be reassembled for data augmentation. This approach can effectively improve the model's generalization ability and allows for more targeted data sampling from the perspective of domain or class, especially for rare classes or domains, thus achieving overall performance stability. In Domain Generalization (DG), sufficient intra-class variability is crucial, a functionality not achievable through traditional image-level supervised losses. Therefore, in DR grading, the diverse diagnostic patterns across different domains encourage the model to learn retinal lesion semantics at the pixel level as extensively as possible, as pixel-level supervised representations are likely to be noisy, which is overlooked in prior works~\cite{che2023towards}. By introducing decoupled retinal semantics for pixel-level alignment alongside severity supervision at the image level, we encourage the model to learn features with adequate intra-class variability while preserving diagnostic patterns.
\par Our contributions can be summarized as: 
(1) We propose decoupling the representation of semantic features from domain features and utilizing a combination of semantic features and features from different domains for feature-level data augmentation. Moreover, to improve the robustness of the disentangled representations, class and domain prototypes are separately interpolated into their corresponding representations;
(2) We design a robust pixel-level alignment loss to align retinal semantics decoupled from representations;
(3) Extensive experiments on comprehensive benchmarks demonstrate the effectiveness of our framework on unseen domains.

\begin{figure}[t]
    \centering
    \includegraphics[width=\linewidth]{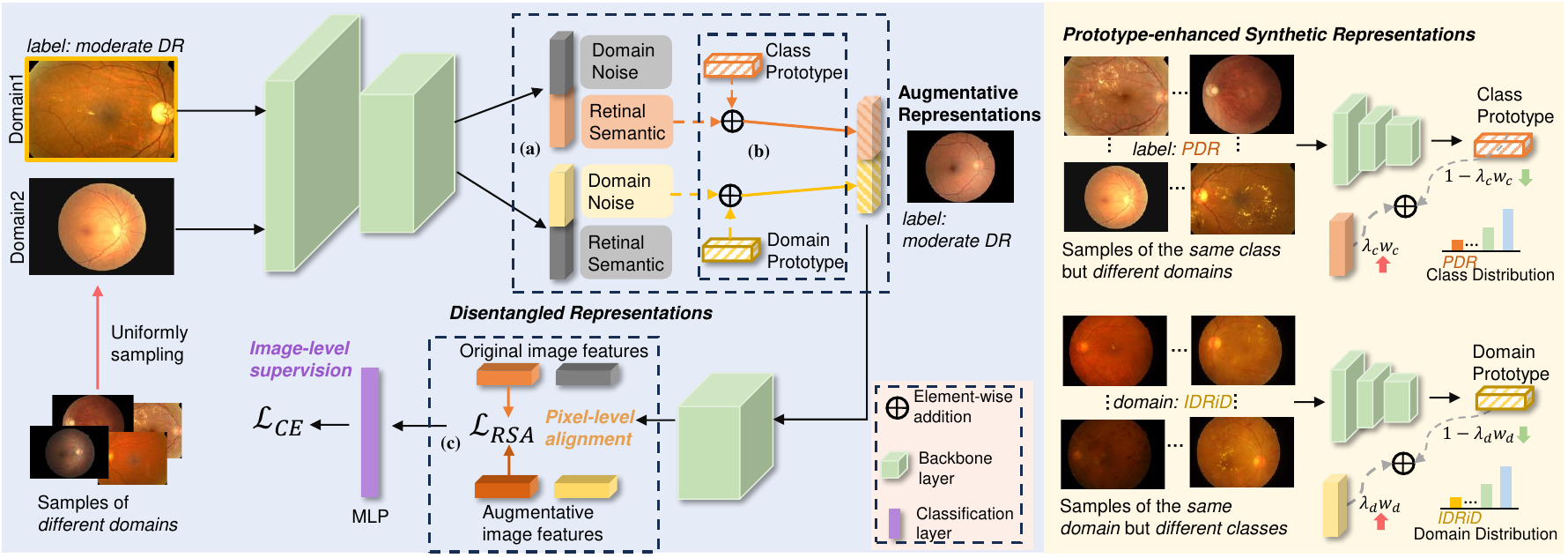}
    \caption{The overview of our proposed method. (a) Representation decoupling and recombination. (b) Representation enhancement with class and domain prototypes. The specific process is shown in the right panel. (c) Robust pixel-level semantic alignment.}
    \label{fig:pipeline}
\end{figure}
\section{Methodology}
To achieve class-unbiased and domain-invariant representations, we propose DE-CO, outlined as shown in Figure~\ref{fig:pipeline}. The key concept is to decouple retinal image representation into class-related retinal semantics and domain noise. 
DECO migrates domain noise to all instances, which offers greater adaptability compared to manually designed image augmentation methods~\cite{shen2020modeling,liu2022understanding}. Additionally, DECO averages the semantic representations of each category and domain noise separately to interpolate the representation of each instance. Finally, we extract retinal semantics from the features for further pixel-level alignment, enabling the model to learn robust class-related retinal features. 

\noindent
\textbf{Task Settings.}
In DG, let $\mathcal{D}=\{D_1, \ldots, D_S\}$ denote the source domains, where each domain \(D_i\) includes data triplets \((x_i, y_i, d)\) representing input fundus image \(x_i\), label \(y_i\), and the domain \(d\), drawn from the domain-specific distribution \(P_i(x, y)\). For each domain \(d\), we define the number of training
examples in each class as \(n_i=\{n_{d,1}, \ldots, n_{d,C}\}\). The goal of DG is to learn a function \(f: \mathcal{X} \rightarrow \mathcal{Y}\) that minimizes the expected loss over an unseen target domain \(D_T\) with its own unique distribution \(P_T(x, y)\). 

\noindent
\textbf{Decoupling and Recombination of Representations.}
DECO constructs augmented instances by combining the semantic representation of one instance with the domain noise of another, thereby recombining pairs of instances to augment instances of diverse styles. Inspired by style transfer~\cite{huang2017arbitrary}, we employ instance normalization~\cite{ulyanov2016instance} to decouple semantic and domain noise. In an exemplar \((x, y, d)\), the hidden representation at the \(l\)-th ($l \textless L$) layer is denoted as \(r = f^{l}(x)\in \mathbb{R}^{C\times H \times W}\), where \(C\), \(H\), and \(W\) signify the channel, height, and width dimensions, respectively. InstanceNorm normalizes it as:
\begin{equation}
    z(r) = \text{InstanceNorm}(r) = \frac{r-\mu(r)}{\sigma(r)}.
\end{equation}
where $\mu(\cdot)$, $\sigma(\cdot)$ are the mean and standard deviation computed across the spatial dimensions of $r$ for each channel and each sample:
\begin{equation}
\small
        \mu(r) = \frac{1}{HW} \sum_{h=1}^{H} \sum_{w=1}^{W} r_{h,w}, \;\;
        \sigma(r) = \sqrt{\frac{1}{HW} \sum_{h=1}^{H} \sum_{w=1}^{W} (r_{h,w} - \mu(r))^2}.
\end{equation}
InstanceNorm occurs in feature space, where the affine parameters can alter the style of the image~\cite{huang2017arbitrary,ulyanov2016instance}. Therefore, we consider the normalized example \( z(r) \) as the semantic representation, representing the retinal-related semantics in retinal images, while \( \mu(r) \) and \( \sigma(r) \) are regarded as domain noise, encompassing factors such as image style, background, ethnicity, and age differences.\\
After decoupling representations, an augmented representation is generated by interchanging semantic representations and domain noise within a pair of examples \((x_i, y_i, d_i)\) and \((x_j, y_j, d_j)\).
\begin{equation}
    r' = \sigma({r}_i) \left(\frac{r_i - \mu(r_i)}{\sigma(r_i)} \right) + \mu(r_j), \;\; \hat{y} = y_i.
\end{equation}
Combining the semantic representation of \( x_i \) with the domain representation of \( x_j \) yields an augmented representation \( r' \) with the category \( y_i \). By decoupling and recombining representations, numerous augmented representations can be generated, encompassing diverse domain noise, particularly making DECO more reliable when faced with domain or class imbalance.

\noindent
\textbf{Prototype-enhanced Synthetic Representations.}
Although instance normalization can effectively distinguish semantic information from domain noise, it is challenging to entirely shield the semantic information of each instance from domain bias. Therefore, to improve the robustness of semantic representations, we leverage the averaged semantic representations from the same class but different domains~\cite{xia2023hgclip}. Considering the diversity of the same class, we balance class and domain invariance by combining corresponding class prototype representations and semantic representations. The class prototype representation \( p_c \) is an invariant representation belonging to class \( c \), independent of the domain, obtained by averaging semantic representations of instances belonging to class \( c \):
\begin{equation}
    p_c = \frac{1}{n_{\star,c}} \sum_{i=1}^{n_{\star,c}} z(r_i) = \frac{1}{n_{\star,c}} \sum_{i=1}^{n_{\star,c}} \frac{r_i - \mu(r_i)}{\sigma(r_i)},\;\;
    \text{where} \;\; n_{\star,c} = \sum_{k=1}^{S}n_{k,c}.\\
\end{equation}
For every instance \((x, y, d)\) where \(y = c\), the prototype-enhanced semantic representation \(\hat{z}(r)\) is derived through the interpolation of \(z(r)\) with the class prototype \(p_c\). Additionally, considering the class imbalance~\cite{xia2023lmpt,hu2023nurvid}, we design a class-aware weight with the interpolation coefficient \(\lambda_c\) to balance the semantic representations and prototypes of minority groups.
\begin{equation}
\centering
    \begin{aligned}
    \hat{z}(r) = \lambda_c w_c z(r) + (1 - \lambda_c w_c) p_c, \;\;  \small{ w_c = \frac{\sum_{c=1}^C \sum_{k=1}^{S}\left(n_{k,c}\right)^{\gamma_c}}{\sum_{k=1}^{S}\left(n_{k,c}\right)^{\gamma_c}}},
    \end{aligned}
\end{equation}
where \(\gamma_c\) is a scale hyper-parameter to provide more flexibility. Similarly, in the ideal scenario, \( \mu(r) \) and \( \sigma(r) \) exclusively contain domain noise representations, yet they may also include class-related representations, such as certain subtle lesions. In this case, we obtain domain prototype representations \( u_d \) and \( v_d \) by averaging examples from different classes within the same domain to alleviate semantic information:
\begin{equation}
    u_d = \frac{1}{n_{d,\star}} \sum_{i=1}^{n_{d,\star}} \mu(r_i), \quad v_d = \frac{1}{n_{d,\star}} \sum_{i=1}^{n_{d,\star}} \sigma(r_i),\;\;
    \text{where} \;\; n_{d,\star} = \sum_{k=1}^{C}n_{d,k}.\\
\end{equation}
Then, we design a weighting scheme tailored to domains with inadequate representation, followed by interpolating the domain noise of each instance with domain prototypes using domain weights and a balancing coefficient:
\begin{equation}
\centering
    \hat{\mu}(r) = \lambda_d w_d \mu(r) + (1 - \lambda_d w_d) u_d, \quad \hat{\sigma}(r) = \lambda_d w_d \sigma(r) + (1 - \lambda_d w_d) v_d,
\end{equation}
\text{where} $w_d = \frac{\sum_{d=1}^S \sum_{k=1}^{C}\left(n_{d,k}\right)^{\gamma_d}}{\sum_{k=1}^{C}\left(n_{d,k}\right)^{\gamma_d}}$. Finally, by replacing the original semantic representation and domain noise, the prototype-enhanced augmented representation:
\begin{equation}
    \hat{r}' = \hat{\sigma}({r}_j) \hat{z}(r_i) + \hat{\mu}(r_j), \;\; \hat{y}' = y_i.
\end{equation}

\noindent
\textbf{Robust Pixel-level Semantic Alignment Loss.}
Image-level supervised losses (such as cross-entropy) have been the most common, capable of learning dense class features. However, sufficient intra-class variation is crucial for domain generalization, with some works~\cite{che2023towards,chen2020simple} proposing to encourage models to learn pixel-level supervision. However, they overlook that the representations for alignment are susceptible to domain interference. Therefore, we introduce decoupled retinal semantics for pixel-level alignment, along with image-level severity supervision, to encourage the model to learn robust class features on the retina while possessing features with sufficient intra-class variation. For each instance \((x, y, d)\), the original features are \(r = f^{L}(x)\), and its augmented features are \(\hat{r}'\). Our training objective combines \(\mathcal{L}_{img}\) and \(\mathcal{L}_{pixel}\) as follows: 
\begin{equation}
    \footnotesize
    \mathcal{L}_{total} = \mathcal{L}_{CE}\left(\text{MLP}(\hat{r}'),y\right) - \alpha \log \frac{\text{exp}(z(\hat{r}')\cdot z(r)/\tau)}{\sum_{k=1}^{2M}\mathds{1}_{[k \neq i]}\text{exp}(z(\hat{r}')\cdot z(r_k)/\tau)},
\end{equation}
where \(z(\cdot)\) represents semantic features, \(\mathds{1}[k \neq i] \in \{0, 1\}\) is an indicator function that equals 1 when \(k \neq i\), \(\tau\) denotes the temperature parameter, \(\alpha\) dynamically controls the task weight and \(M\) is the size of the randomly sampled mini-batch.

\begin{table}[t]
\centering
\footnotesize
\caption{Comparison with state-of-the-art approaches under the DG test.}
\label{tab:dg}
\resizebox{\linewidth}{!}{
\begin{tabular}{l|ccc|ccc|ccc|ccc|ccc|ccc|ccc}
\toprule
Target & \multicolumn{3}{c|}{APTOS} & \multicolumn{3}{c|}{DeepDR} & \multicolumn{3}{c|}{FGADR} & \multicolumn{3}{c|}{IDRID} & \multicolumn{3}{c|}{Messidor} & \multicolumn{3}{c|}{RLDR} & \multicolumn{3}{c}{Average} \\ 
\midrule Metrics & AUC & ACC & F1 & AUC & ACC & F1 & AUC & ACC & F1 & AUC & ACC & F1 & AUC & ACC & F1 & AUC & ACC & F1 & AUC & ACC & F1 \\ \midrule
ERM & 75.0 & 44.4 &38.9 &77.0& 39.5& 34.3& 66.2& 32.0& 27.1& 82.3 &50.0& 44.1& 79.1& 60.7& 43.4& 75.9& 36.5& 35.7& 75.9& 43.8& 37.3 \\
Mixup~\cite{zhang2018mixup} & 75.3& 62.6& 43.2& 75.3& 29.0& 25.2& 66.7& 42.3& 32.3& 78.8& 39.0& 27.6& 76.7& 54.7& 32.6& 76.9& 43.6& 37.7& 75.0& 45.2& 33.1 \\
MixStyle~\cite{zhou2020domain} & 79.0& 65.8& 39.9& 76.9& 32.9& 27.9& 71.2& 35.8& 22.7& 83.0& 51.4& 39.2& 75.2& 62.2& 36.5& 75.5& 41.1& 31.4& 76.8& 48.2& 32.9 \\
GREEN~\cite{liu2020green}& 75.1& 53.8& 38.9& 76.4& 28.1& 24.9& 69.5& 41.3& 31.5& 79.9& 41.3& 32.2& 75.8& 52.0& 36.8& 74.8& 34.0& 34.4& 75.3& 41.8& 33.1 \\
CABNet~\cite{he2020cabnet} & 75.8& 55.5& 39.4 &75.2 &42.7 &31.8 &73.2 &43.7 &34.8& 79.2& 44.8 &37.3& 74.2& 56.1 &34.1& 75.8 &37.0 &35.6 &75.6 &46.6 &35.5 \\
DDAIG~\cite{zhou2020deep} & 78.0& 67.1 &41.0 &75.6 &37.6 &32.2 &73.6 &42.0& 33.8 &82.1 &37.4 &27.0 &76.6 &58.4 &35.3 &75.6 &36.1 &27.7& 76.9& 46.4 &32.8 \\
ATS~\cite{yang2021adversarial} & 77.1 &56.9 &38.3 &79.4 &36.1 &31.6 &74.7 &46.7 &33.4 &83.0 &41.5 &34.9 &77.2 &64.7 &35.8 &76.5 &37.4& 34.9& 78.0&47.2& 34.8 \\
Fishr~\cite{rame2022fishr} & 79.2& 66.6 &43.4 &81.1 &48.1&34.4& 73.3&44.4 &34.4& 82.7&40.3 &27.6 &76.4 &\textbf{65.1} &41.1 &77.4 &36.8 &34.7 &78.4 &50.2 &35.9 \\
MDLT~\cite{yang2022multi}& 77.3 &57.2& 41.5& 80.0 &39.5 &36.2& 74.1 &45.7& 29.0 &81.5 &44.2 &35.4&75.4& 58.9 &36.9 &75.7 &37.6 &35.0 &77.3& 47.2 &35.7 \\
DRGen~\cite{atwany2022drgen} & 79.9 &58.1& 40.2& 83.0& 38.7& 34.1& 69.4& 41.7& 24.7& 84.7& \textbf{44.6}& 37.4& 79.0& 60.1& 40.5& 79.5& 43.1& 37.0& 79.3& 47.7& 37.3 \\ 
VAE-DG~\cite{chokuwa2023generalizing} & 79.5 & 66.7 & 45.5 & 82.9 & 38.1 & 34.3 & 75.8 & 44.8 & 36.4 & 83.6 & 41.7 & 35.6 & 78.7 & 59.5 & 40.2 & 78.8 & 41.3 & 35.9 & 79.9 & 48.7 & 38.0 \\ 
GDRNet~\cite{che2023towards} &79.9 &66.8 &46.0&84.7 &53.1 &45.3& 80.8 &45.3 &39.4 &84.0 &40.3& 35.9 &\textbf{83.2} &63.4 &50.9 &82.9 &45.8 &43.5& 82.6 &52.5 &43.5 \\ \midrule
DECO(Ours) & \textbf{81.0} & \textbf{72.4} & \textbf{56.6} & \textbf{86.3} & \textbf{57.3} & \textbf{52.6} & \textbf{81.3} & \textbf{49.1} & \textbf{43.5} & \textbf{87.0} & \textbf{44.6} & \textbf{37.4} & 82.9 & \textbf{65.1} & \textbf{52.7} & \textbf{83.5} & \textbf{48.0} & \textbf{46.3} & \textbf{83.7} & \textbf{56.1} & \textbf{48.2} \\ \bottomrule
\end{tabular}}
\label{tab:result1}
\vspace{-2em}
\end{table}

\section{Experiments}
\textbf{Datasets and Metrics.}
We conduct a comprehensive evaluation of our method on GDRBench~\cite{che2023towards}, which involves two generalization ability evaluation settings and eight popular public datasets. First, GDRBench adopts the classic leave-one-domain-out protocol (DG test), which requires leaving one domain for evaluation and training models on the rest. It involves six datasets, including DeepDR~\cite{liu2022deepdrid}, Messidor~\cite{abramoff2016improved}, IDRID~\cite{porwal2018indian}, APTOS~\cite{karthick2019aptos}, FGADR~\cite{zhou2020benchmark}, and RLDR~\cite{wei2021learn}. Additionally, it incorporates an extreme single-domain generalization setup (ESDG test), following the train-on-single-domain protocol on the above datasets for training, but with two extra large-scale datasets, DDR~\cite{li2019diagnostic} and EyePACS~\cite{emma2015eyepacs} for evaluation.

For evaluation, we report three critical metrics, namely accuracy (ACC), the area under the ROC curve (AUC), and macro F1-score (F1). \\ \noindent
\textbf{Implementation Details.}
We used ResNet-50~\cite{he2016deep} pre-trained on ImageNet~\cite{deng2009imagenet} as the backbone and a fully connected layer as the linear classifier. We use the AdamW optimizer with a learning rate of $5\times 10^{-4}$, weight decay of $10^{-4}$ and a batch size of 32. The model is trained for 100 epochs. In our experiments, we apply cross-validation to tune all hyperparameters with grid search. 

\noindent
\textbf{Baselines.} Following~\cite{che2023towards}, our comparative algorithms, besides the naive Empirical Risk Minimization (ERM), are mainly divided into three categories: conventional DR classification algorithms~\cite{he2020cabnet,liu2020green}, domain generalization techniques~\cite{atwany2022drgen,zhang2018mixup,zhou2020deep,zhou2020domain,yang2021adversarial,chokuwa2023generalizing}, and feature representation methods~\cite{yang2022multi,rame2022fishr}.
\begin{table}[t]
\centering
\footnotesize
\caption{Comparison with state-of-the-art approaches under the ESDG test.}
\label{tab:esdg}
\resizebox{\linewidth}{!}{
\begin{tabular}{l|ccc|ccc|ccc|ccc|ccc|ccc|ccc}
\midrule
Source & \multicolumn{3}{c|}{APTOS} & \multicolumn{3}{c|}{DeepDR} & \multicolumn{3}{c|}{FGADR} & \multicolumn{3}{c|}{IDRID} & \multicolumn{3}{c|}{Messidor} & \multicolumn{3}{c|}{RLDR} & \multicolumn{3}{c}{Average} \\ 
\midrule Metrics & AUC & ACC & F1 & AUC & ACC & F1 & AUC & ACC & F1 & AUC & ACC & F1 & AUC & ACC & F1 & AUC & ACC & F1 & AUC & ACC & F1 \\ \midrule
ERM & 66.4& 53.2& 31.6 &70.7 &47.3 &31.2 &55.3& 5.6 &7.1 &69.6 &56.5 &33.9 &70.6 &51.3 &33.7& 70.1 &27.3& 26.4 &67.1& 40.2& 27.3 \\
Mixup~\cite{zhang2018mixup} & 65.5& 49.4& 30.2& 70.7& 49.7& 33.3 &58.8 &5.8& 7.4& 70.2 &64.0& 32.6& 71.5& 63.0 &32.6 &72.9 &27.7 &27.0 &68.3 &43.3 &27.2 \\
MixStyle~\cite{zhou2020domain} & 62.0& 48.8& 25.0& 53.3 &32.0 &14.6 &51.0& 7.0&7.9& 53.0 &53.5 &19.4 &51.4 &57.6 &16.8& 53.5& 18.3 &6.4 &54.0 &36.2 &15.0  \\
GREEN~\cite{liu2020green}& 67.5& 52.6& 33.3& 71.2 &44.6 &31.1 &58.1& 5.7& 6.9 &68.5& 60.7 &33.0& 71.3 &54.5& 33.1 &71.0 &31.9& 27.8&67.9& 41.7& 27.5 \\
CABNet~\cite{he2020cabnet} & 67.3& 52.2& 30.8& 70.0& 55.4 &32.0 &57.1& 6.1 &7.5 &67.4 &62.7 &31.7& 72.3& 63.8& 35.3 &75.2 &23.0& 25.4& 68.2 &43.8& 27.2 \\
DDAIG~\cite{zhou2020deep} & 67.4 &48.7 &31.6 &73.2 &38.5 &29.7& 59.9& 5.0 &5.5& 70.2 &60.2 &33.4 &73.5 &69.1 &35.6& 74.4 &25.4 &23.5& 69.8& 41.2 &26.7 \\
ATS~\cite{yang2021adversarial} & 68.8 &51.7 &32.4& 72.7 &52.4& 33.5 &60.3 &5.3& 5.7& 69.1& 66.6 &30.6& 73.4 &64.8 &32.4 &75.0& 24.2& 23.9 &69.9& 44.2 &26.4 \\
Fishr~\cite{rame2022fishr} & 64.5 &61.7& 31.0 &72.1& \textbf{61.0}& 30.1 &56.3 &6.0 &7.2& 71.8 &48.0 &30.6 &74.3 &52.0 &33.8 &78.6 &19.3& 21.3& 69.6 &41.3 &25.7 \\
MDLT~\cite{yang2022multi}& 67.6& 53.3& 32.4& 73.1 &50.2 &33.7& 57.1& 7.1 &7.8& 71.9&61.7& 32.4 &73.4 &58.9& 34.1 &76.6& 29.0 &30.0 &70.0 &43.4& 28.4 \\
DRGen~\cite{atwany2022drgen} & 69.4& \textbf{60.7}& 35.7 &\textbf{78.5}& 39.4 &31.6& 59.8 &6.8 &8.4 &70.8 &67.7 &30.6 &77.0 &64.5 &37.4 &78.9 &19.0 &21.2 &72.4 &43.0& 27.5 \\
VAE-DG~\cite{chokuwa2023generalizing} & 67.7 & 51.9 & 32.3 & 75.4 & 53.6 & 33.4 & 60.2 & 5.1 & 5.6 & 72.4 & 63.0 & 32.3 & 75.8 & 59.9 & 35.7 & 78.9 & 23.6 & 24.0 & 71.7 & 42.9 & 27.2 \\ 
GDRNet~\cite{che2023towards} & 69.8 &52.8 &35.2 &76.1 &40.0 &35.0 &63.7& 7.5& 9.2& 72.9 &70.0& 35.1 &78.1 &65.7 &40.5 &79.7 &44.3 &37.9 &73.4& 46.7& 32.2 \\ \midrule
DECO(Ours) & \textbf{70.6} & 59.7 & \textbf{36.4} & 78.2 & 40.3 & \textbf{35.8} & \textbf{65.5} & \textbf{9.9} & \textbf{11.7} & \textbf{74.2} & \textbf{74.8} & \textbf{38.7} & \textbf{79.8} & \textbf{70.1} & \textbf{44.9} & \textbf{81.1} & \textbf{49.3} & \textbf{40.8} & \textbf{75.0} & \textbf{50.7} & \textbf{34.7} \\ \bottomrule
\end{tabular}}
\label{tab:result2}
\vspace{-1em}
\end{table}

\noindent\textbf{Comparison with SoTA methods under the DG test.}
Table \ref{tab:result1} demonstrates that our approach consistently outperforms previous state-of-the-art methods across all datasets. Furthermore, notable improvements are observed, particularly in domains with limited representation such as IDRiD, indicating the efficacy of our method in clinical applications. The decoupled feature enhancement and semantic alignment exhibit outstanding performance in scenarios with limited samples. Traditional DR classification methods, which do not account for domain shifts, exhibit weak generalization. In contrast, domain generalization and feature representation methods show improvements over the baseline, attributed to the strategic design addressing domain shifts. In summary, DECO significantly outperforms these SoTA methods. The decoupling representations effectively achieve feature-level data augmentation, while the incorporation of class or domain prototypes further enhances the robustness of augmented data. Additionally, robust pixel-level semantic alignment enables the model to learn precise intra-class variations, thereby improving model generalization. 

\noindent
\textbf{Generalization from a single source domain.}
We further comprehensively evaluate the generalization performance by training on single-domain datasets and testing on large-scale unseen datasets. The results, as shown in Table \ref{tab:result2}, indicate a significant performance decrease for all methods, highlighting the challenge of the task. Despite the decreased utilization of feature-level data augmentation in the ESDG test, DECO still outperforms other methods on most datasets due to its enhanced diversity within the same class samples and the learned robust domain-invariant representations. 

\noindent
\textbf{Ablation study on the components.} 
In Table \ref{tab:aba}, we conduct an ablation analysis on the three components. Firstly, the Augmented Disentangled Representations (ADR) significantly improve the model's generalization ability. Subsequently, under the joint influence of class and domain prototypes, more robust augmented representations further optimize the model. Finally, robust pixel-level semantic alignment (RSALoss) of the disentangled semantic representations enhances the model's generalization by improving robust intra-class diversity. 

\noindent
\textbf{Further analysis of augmented methods and $\mathcal{L}_{pixel}$.}
To further analyze the effectiveness of ADR and RSALoss, the primary components of our approach, we compared them with state-of-the-art techniques in the same category. For the augmentation methods, the comparative methods included the standard DG augmentation \cite{yang2022multi}, the default augmentation strategy in DRGen \cite{atwany2022drgen}, visual transformation and image degradation \cite{che2023towards}. Our proposed disentangled representation augmentation significantly outperforms the other methods. Regarding semantic alignment loss, we compared models without semantic alignment loss and with DahLoss \cite{che2023towards}. Due to the improved robustness of alignment features with decoupled semantic representations, RSALoss demonstrates significant advantages.
\\ \noindent
\begin{table}[t]
\centering
\scriptsize
\caption{Ablation studies on proposed components under the DG test.}
\label{tab:abla}
\begin{tabular}{ccc|cccccc|c}
\toprule
ADR & Prototype & $\mathcal{L}_{pixel}$ & APTOS & DeepDR & FGADR & IDRID & Messidor & RLDR & Average \\  \midrule
- & - & - & 75.26 & 77.21 & 66.35 & 82.44 & 79.29 & 75.95 & 76.08 \\
\ding{51} & - & - & 78.91 & 84.20 & 78.17 & 84.33 & 80.95 & 81.56 & 81.35 \\
- & - & \ding{51} & 79.53 & 83.16 & 76.99 & 84.05 & 81.02 & 81.64 & 81.07 \\
\ding{51} & \ding{51} & - & 80.15 & 85.13 & 79.56 & 86.37 & 82.11 & 82.84 & 82.69 \\
\ding{51} & - & \ding{51} & 80.28 & 85.40 & 79.83 & 86.30 & 82.43 & 82.96 & 82.87 \\
\ding{51} & \ding{51} & \ding{51} & 81.03 & 86.29 & 81.26 & 86.99 & 82.94 & 83.50 & 83.67 \\
\bottomrule
\end{tabular}
\label{tab:aba}
\end{table}

\begin{figure}[t]
    \centering
    \includegraphics[width=0.85\linewidth]{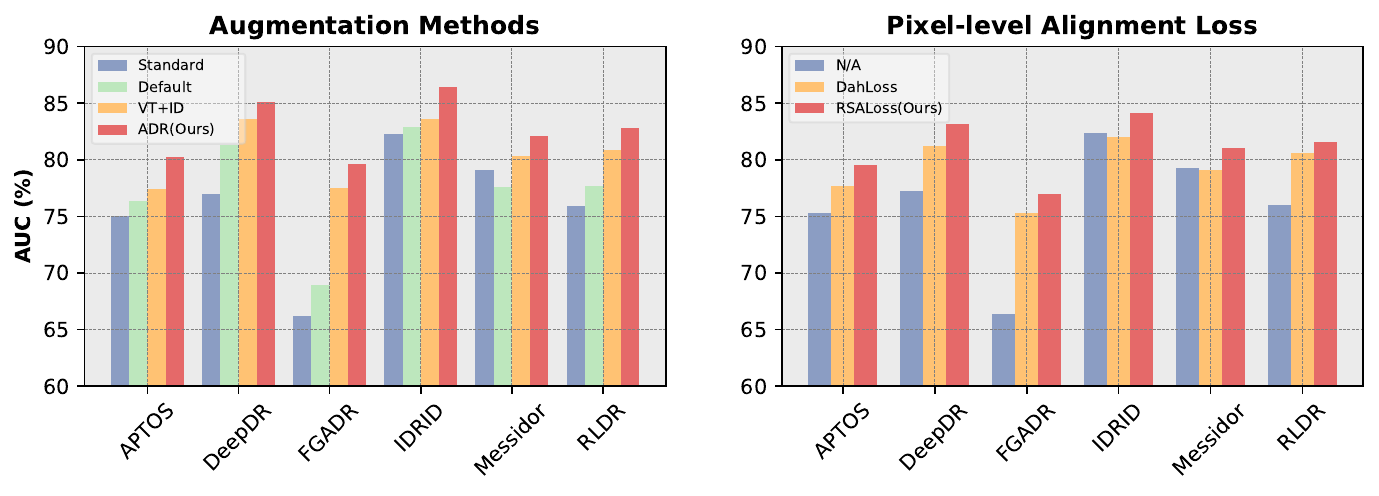}
    \vspace{-1em}
    \caption{Analysis of augmentation methods and $\mathcal{L}_{pixel}$ under the DG test.}
    \label{fig:fur}
\end{figure}
\vspace{-1.5em}
\section{Conclusion}
In this work, we mitigate the performance gap faced by DR grading models due to domain shifts with disentangled representations. We innovatively propose a method to combine semantic representations relevant to DR classes with domain noise from another domain. It enhances the model's generalization to unseen domains. Then, to improve the robustness of decoupled representations, we derive prototypes by averaging class and domain representations and designing data-aware coefficients to adjust focus on different classes/domains, facilitating interpolation between semantic representations and prototypes. Finally, we introduce a robust pixel-level semantic alignment loss to simultaneously learn densely packed class features while preserving inter-class diversity.

%
%
%
\bibliographystyle{splncs04}
\bibliography{main}

\newpage
\textbf{\centering Appendix for "Generalizing to Unseen Domains in Diabetic Retinopathy with Disentangled Representations"}

\begin{figure}[h]
    \centering
    \includegraphics[width=0.8\linewidth]{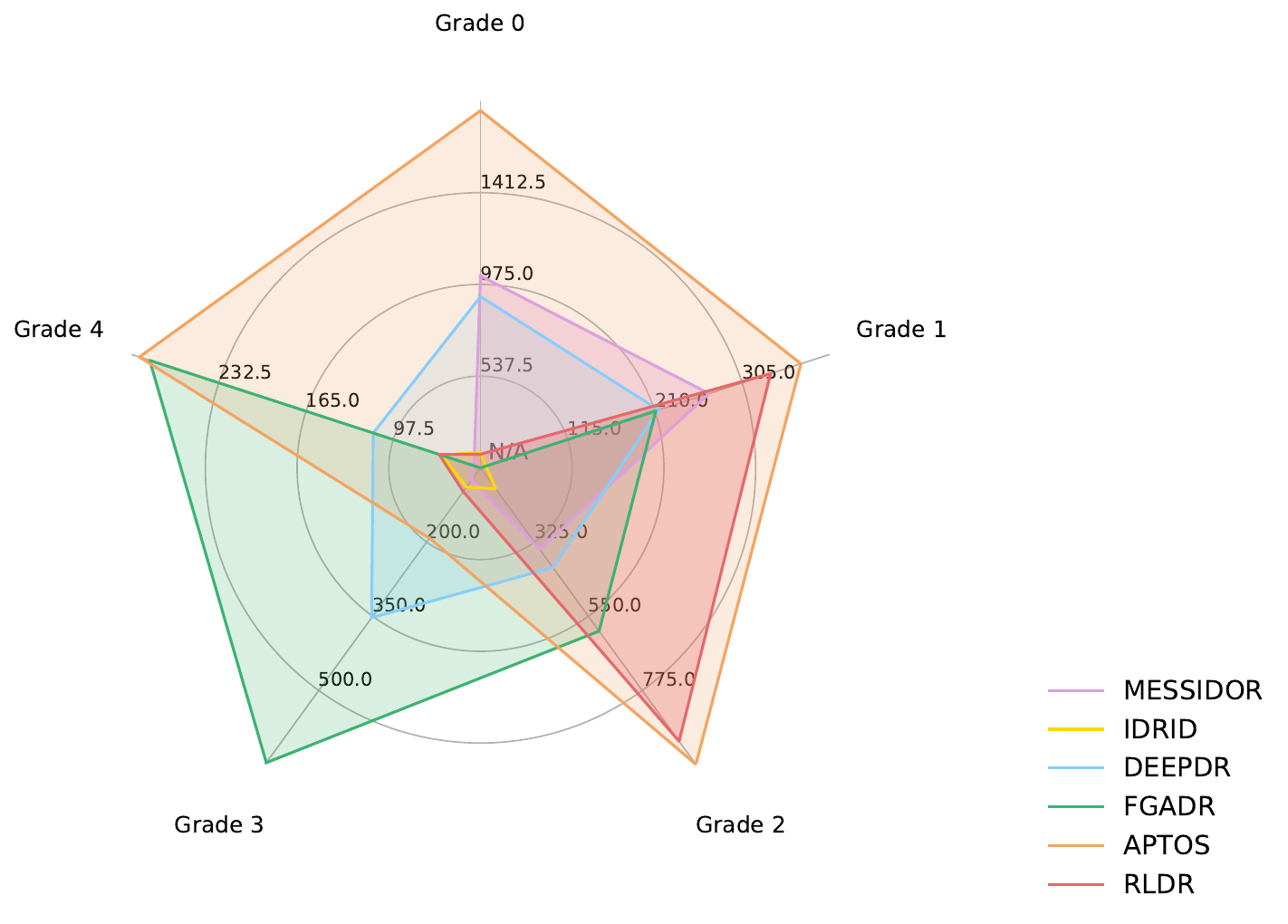}
    \caption{Data distribution for each category in 6 datasets.}
\end{figure}
\begin{figure}[htbp]
    \centering
    \includegraphics[width=0.5\linewidth]{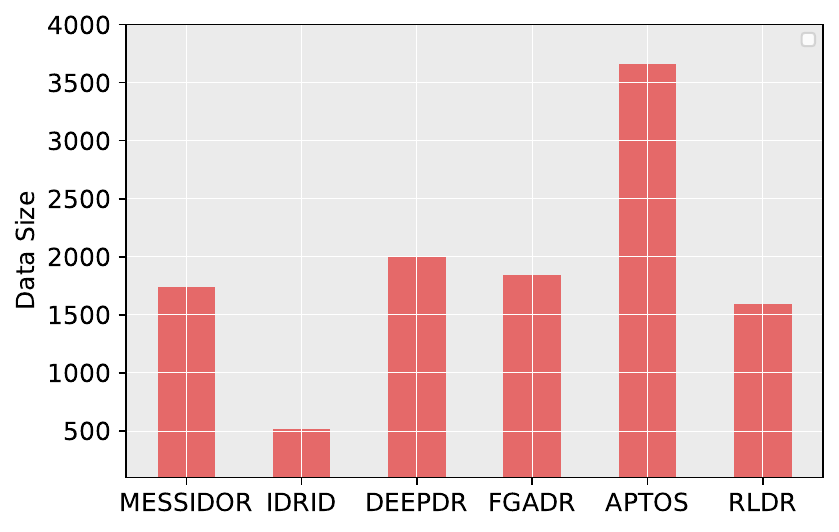}
    \caption{Distribution of individual datasets.}
\end{figure}

\begin{table}[h]
\small
    \centering
        \caption{Hyperparameters for experiments. $^\dag$ denotes we adopt a warm start strategy of running vanilla ERM for the first few epochs to ensure reliable disentanglement. Interpolation coefficient $\lambda_c \sim \;\text{Beta}(\alpha_c, \alpha_c)$ and $\lambda_d \sim \;\text{Beta}(\alpha_d, \alpha_d).$}
        \resizebox{\linewidth}{!}{
    \begin{tabular}{l|ccc}
        \toprule
        Hyperparameters & DECO (DG) & DECO (ESDG) & Comparison methods  \\ \midrule
        Epochs & 100 & 100 & 100 \\
        Learning Rate & $5\times 10^{-4}$ & $5\times 10^{-4}$ & $5\times 10^{-4}$ \\ 
        Weight Decay & $10^{-4}$ & $10^{-4}$ & $10^{-4}$ \\
        Batch Size & 32 & 32 & 32 \\
        Warm Start Epochs$^\dag$ & 30 & 35 & - \\
        Class Prototype Mixup Parameter $\alpha_c$ & 0.5 & 0.55 & - \\
        Domain Prototype Mixup Parameter $\alpha_d$ & 0.5 & 0.4 & - \\
        \(\gamma_c\) & 0.2 & 0.2 & - \\
        \(\gamma_d\) & 0.2 & - & - \\
        \bottomrule
    \end{tabular}}
\end{table}

\end{document}